\documentclass{article}

\usepackage{arxiv}

\usepackage[utf8]{inputenc} 
\usepackage[T1]{fontenc}    
\usepackage{hyperref}       
\usepackage{url}            
\usepackage{booktabs}       
\usepackage{amsfonts}       
\usepackage{nicefrac}       
\usepackage{microtype}      
\usepackage{lipsum}
\usepackage{graphicx}
\graphicspath{ {./images/} }
\usepackage{paralist}
\usepackage{xcolor}
\usepackage{amsmath}

\title{Measuring LLM Sensitivity in Transformer-based Tabular Data Synthesis}

\author{
 Maria F. Davila R.\thanks{Corresponding Author: maria.fernanda.davila.restrepo@uni-oldenburg.de} \\
  Computer Science Department\\
  Carl von Ossietzky University\\
  Oldenburg, Germany \\
   \And
 Azizjon Turaev\\
  Computer Science Department\\
  Carl von Ossietzky University\\
  Oldenburg, Germany \\
  \And
 Wolfram Wingerath \\
  Computer Science Department\\
  Carl von Ossietzky University\\
  Oldenburg, Germany \\
}

\begin{document}
\maketitle
\begin{abstract}
Synthetic tabular data is used for privacy-preserving data sharing and data-driven model development. Its effectiveness, however, depends heavily on the used Tabular Data Synthesis (TDS) tool. Recent studies have shown that Transformer-based models outperform other state-of-the-art models such as Generative Adversarial Networks (GANs) and Diffusion models in terms of data quality. However, Transformer-based models also come with high computational costs, making them sometimes unfeasible for end users with prosumer hardware. This study presents a sensitivity assessment on how the choice of hyperparameters, such as number of layers or hidden dimension
affects the quality of the resultant synthetic data and the computational performance. It is performed across two tools, GReaT and REaLTabFormer, evaluating 10 model setups that vary in architecture type and depth. We assess the sensitivity on three dimensions: runtime, machine learning (ML) utility, and similarity to real data distributions. Experiments were conducted on four real-world datasets. 
Our findings reveal that runtime is proportional to the number of hyperparameters, with shallower configurations completing faster. GReaT consistently achieves lower runtimes than REaLTabFormer, and only on the largest dataset they have comparable runtime. For small datasets, both tools achieve synthetic data with high utility and optimal similarity, but on larger datasets only REaLTabFormer sustains strong utility and similarity. As a result, REaLTabFormer with lightweight LLMs provides the best balance, since it preserves data quality while reducing computational requirements. Nonetheless, its runtime remains higher than that of GReaT and other TDS tools, suggesting that efficiency gains are possible but only up to a certain level.
\end{abstract}

\section{Introduction}

Synthetic data is used when access to real data is limited due to privacy concerns or scarcity~\cite{Figueira, Hernandez, Fan}. It is popular across domains such as images, text, and tabular data. In this study, we focus on Tabular Data Synthesis (TDS) and define tabular data as consisting of one or more tables, where rows represent records and columns represent features~\cite{Codd70, MolinaDBBook2009}.

Synthetic data is only useful if TDS tools are capable of generating high-quality data. Unlike synthetic images, where quality can be assessed visually, the quality of tabular data is harder to measure. It depends heavily on the type of the data and its intended use. As discussed by Goodfellow et al.~\cite{goodfellow_deep_2016}, different applications have different requirements for fidelity, diversity, and privacy. In simple terms, high-quality synthetic data should preserve the key patterns and distributions of the original dataset without duplicating existing records.

In earlier work~\cite{davila_r_navigating_2024}, we introduced a benchmark to evaluate the synthetic data quality generated by TDS tools and their underlying models across six critical dimensions: handling of class imbalance, dataset augmentation, imputation of missing values, privacy preservation, ML utility, and computational performance. These dimensions were chosen based on the main use cases for synthetic data and the \textbf{\textit{privacy versus utility trade-off}}~\cite{park_data_2018}. 

Data utility refers to how well synthetic data can serve the same purpose as the real dataset. For example, if the original data is used to train a classification model, the utility of the synthetic dataset is its ability to train a model with similar predictive performance. However, generating synthetic data that protects privacy often comes at the cost of reduced utility~\cite{ge_kamino_2021}. Simple anonymization methods, such as noise injection, can break the patterns in the data~\cite{yang_tabular_2024}.

\begin{figure}
    \centering
    \includegraphics[width=1\linewidth]{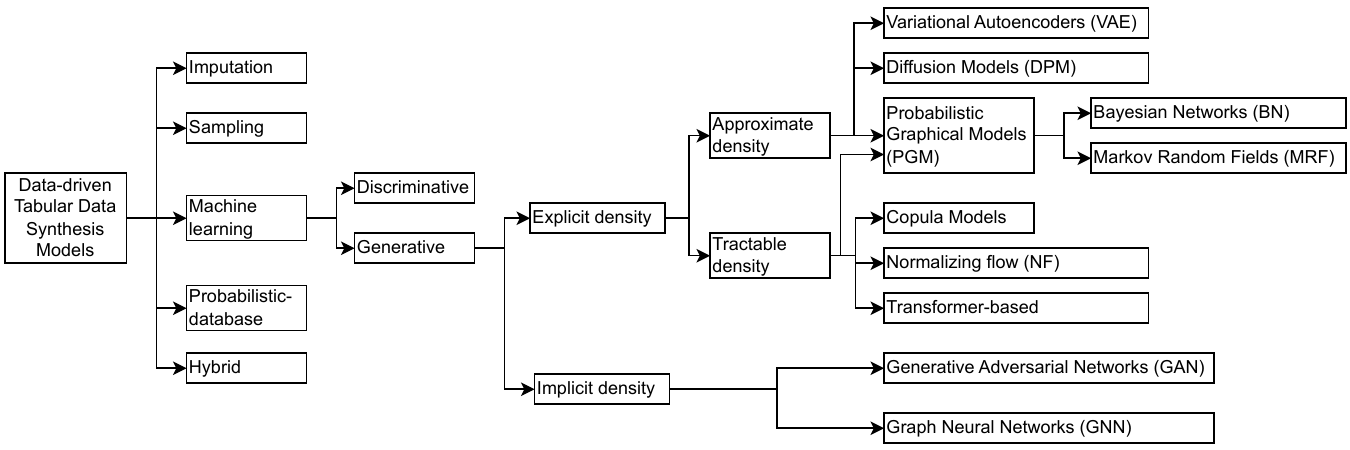}
    \caption{Taxonomy of data-driven TDS models (illustration taken from~\cite[Fig. 1]{davila_r_navigating_2024}).}
    \label{fig:taxonomy}
\end{figure}

The evaluated tools were selected based on the taxonomy shown in Figure~\ref{fig:taxonomy}. Our results showed that Transformer-based models outperformed other state-of-the-art models in five of the six evaluation dimensions. The only dimension where they underperformed was computational performance, which included CPU usage, GPU usage, memory consumption, and total runtime during synthetic data generation.

These strong results could position Transformer-based tools as the default choice for practitioners. However, their high computational demands limit their accessibility for users with prosumer hardware. Therefore, our motivation is to assess whether it is possible to improve the computational performance of Transformer-based TDS tools, without compromising synthetic data quality. This paper explores whether the quality of synthetic data is heavily influenced by the choice of the language model, and if similar quality can be achieved using models with lower resource requirements.

Our contribution is a sensitivity assessment of how the choice of large language model (LLM) affects the quality and performance of synthetic tabular data generation. We evaluate sensitivity along three key dimensions: runtime, ML utility, and data similarity. The LLM parameters include the model type (such as GPT2 and GPT-NeoX), number of layers, and the TDS tool (REaLTabFormer and GReaT).

The remainder of this paper is organized as follows. Section~\ref{sec:related} reviews related work and justifies our choice of tools and evaluation metrics. Section~\ref{sec:method} describes the experimental setup. Section~\ref{sec:results} presents our findings across the three sensitivity dimensions. Finally, Section~\ref{sec:conclusion} concludes the paper and outlines future work.

\section{Related Work}
\label{sec:related}
This section first gives the theoretical basis to understand Transformer-based TDS models, and further describes the tools used in this assessment, GReaT and REaLTabFormer. In this paper, models refer to the underlying architectures used to generate synthetic tabular data. Models define the theoretical and mathematical principles governing synthetic data generation. Each model is based on a core algorithm, which provides the computational framework for learning and generating data. On the other hand, tools refer to specific implementations of these models. Tools transform theoretical models into practical applications by integrating them into software packages. Finally, this section also presents an introduction to synthetic tabular data evaluation, in order to present the metrics used in the assessment. 

\subsection{Transformer-based TDS Models}
Based on the TDS model taxonomy in Figure~\ref{fig:taxonomy}, Transformer-based TDS models are part of the \textit{Machine Learning}, \textit{generative}, \textit{explicit}, and \textit{tractable density} categories. In this context, \textbf{ML} refers to TDS models that learn patterns directly from data rather than relying on predefined rules. \textbf{Generative} means that these models are designed not only to discriminate between classes but to generate new synthetic data that follows the same distribution as the original. \textbf{Explicit} means that the models specify a concrete probability distribution for the data, in contrast to implicit approaches such as GANs. Finally, \textbf{tractable density} means that the probability of any given data point under the model can be computed exactly~\cite{foster_generative_2019, tomczak_deep_2022}. 

Unlike recurrent neural networks (RNNs) and long short-term memory networks (LSTMs) that process data step by step, Transformers use self-attention to consider entire sequences at once, which greatly improves modeling of long-range dependencies. The \textbf{\textit{self-attention mechanism}} ~\cite{vaswani_attention_2017} can determine what words in a sentence, or columns in a dataset, are the most important to understand it. It operates by first transforming each input element into three vectors: 
\begin{compactitem}
    \item \textbf{Query} (\(\mathbf{Q}\)): The Query represents what the current element is searching for in other elements.
    \item \textbf{Key} (\(\mathbf{K}\)): The Key represents the content each element provides. 
    \item \textbf{Value} (\(\mathbf{V}\)): The Value holds the actual representation to be retrieved.
\end{compactitem}
The attention scores are obtained by multiplying Queries with Keys, producing an \textbf{attention matrix} \(\mathbf{QK}^T\) that measures how relevant the elements are to each other. These scores are multiplied by the Value vectors, and their sum gives the updated representation of each input, composed by the original value and its importance.

LLMs build directly on the Transformer architecture described above, extending its self-attention mechanism to very large scales. An LLM is a neural network of stacked Transformer blocks, where the parameters are the numerical weights that the model learns during training, which are after multiplied by the input token to calculate the $Q$, $K$ and $V$ for each token. The total number of parameters defines the model’s ability to capture patterns in data ~\cite{zhao2025surveylargelanguagemodels}. This process is known as autoregressive data generation, hence many references categorize this group of models as autoregressive, but we refer to them as Transformer-based because, in TDS, all the relevant tools leverage the Transformer architecture.

\begin{figure}[h]
    \centering
    \includegraphics[width=0.9\linewidth]{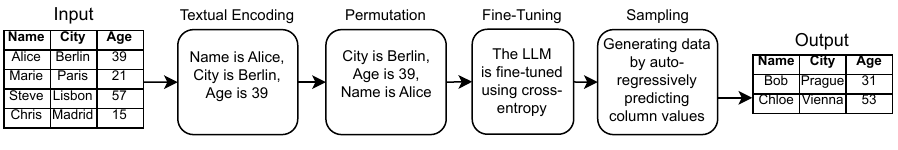}
    \caption{Overview of the synthetic data generation process of the Transformer-based tool, GReaT.}
    \label{fig:great}
\end{figure}

Currently, there are a few transformer-based TDS tools, with GReaT~\cite{GReaT}, and REaLTabFormer  ~\cite{REaLTabFormer}, being the most popular examples. They both use pre-trained LLMs to learn the patterns in the dataset. GReaT transforms tabular data into natural language sentences by doing \textit{textual encoding} of the dataset, which means it turns every row into a sentence using a subject-predicate-object structure. Then, each row is permuted, which means the order of the columns is shuffled to remove artificial order. After, the model is fine-tuned using the input dataset. This means that the pretrained LLM is further trained on the textualized tabular data, adjusting its parameters to better capture the statistical properties and dependencies within the dataset. After fine-tuning is complete, GReaT is able to generate new samples by autoregressively predicting column values. Autoregressively predicting means generating each token sequentially by conditioning on the previously generated tokens. Given a set of column-value pairs, the model samples the next token from the learned conditional probability distribution. This is shown in Figure \ref{fig:great}.

\begin{figure}[h]
    \centering
    \includegraphics[width=0.6\linewidth]{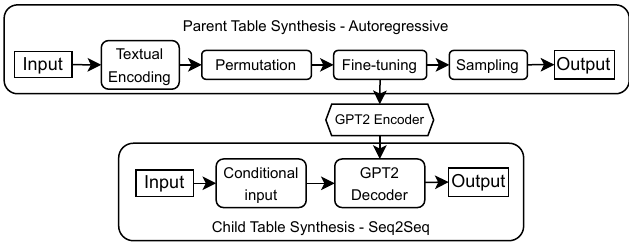}
    \caption{Overview of the REalTabFormer tool, using two models to generate data for relational tables.}
    \label{fig:realtabformer}
\end{figure}

On the other hand, REaLTabFormer ~\cite{REaLTabFormer} uses a two-step approach that first generates a parent table using an autoregressive LLM model, similar to GReaT, and then introduces a second model to generate child tables that maintain relational dependencies.  
This process is shown in Figure \ref{fig:realtabformer}, where the parent table generation is similar to that of GReaT, but now there is also a step for child-table generation. Here, the child-table generation follows a sequence-to-sequence (Seq2Seq) approach, where the pretrained model from the parent table acts as an encoder for the relational dataset. First, the generated parent table is used to condition the child table generation. This is then fed into a transformer decoder, which autoregressively predicts the child table values while ensuring relational constraints are maintained.

\subsection{Synthetic Tabular Data Evaluation}
\label{sec:related_evaluation}
The evaluation of synthetic tabular data remains challenging, as no universally accepted metrics exist, which measure the quality of the synthetic data generated with TDS tools~\cite{chundawat_universal_2024}. We argue that the reason for this is that the choice of evaluation method largely depends on the intended use of the data ~\cite{davila_r_navigating_2024}, which makes a single metric insufficient to capture all relevant aspects.  

Many metrics have been proposed ~\cite{yang_structured_2023,apellaniz_synthetic_2024,lautrup_syntheval_2024}, each targeting specific properties such as correlation structure, distributional similarity, or machine learning utility. One early effort to standardize evaluation is TabSynDex ~\cite{chundawat_universal_2024}, which aggregates several measures into a single score.  

The Synthetic Data Vault (SDV) ~\cite{SDV} also provides the open-source library \texttt{sdmetrics} ~\cite{noauthor_sdmetrics_2024}, implementing a broad set of metrics, including Correlation Similarity, Key Uniqueness, Missing Value Similarity, and Sequence Length Similarity, among many others. In this work, we use a selection of metrics from \texttt{sdmetrics} and extend them to support our sensitivity analysis, as described in Section~\ref{sec:method}.

\section{Methodology for Evaluating TDS Tool Sensitivity to Hyperparameters}
\label{sec:method}

This section first describes how we selected the LLMs included in the assessment. Then, it presents the method we developed to calculate the size of the LLM for each experiment, which is our approach to make the experiments comparable between LLM families. Finally it presents the experimental setup in terms of the datasets used for evaluation, their pre-processing, and the hardware on which the experiments were conducted.

\subsection{LLM Selection}
The LLMs currently available can be divided into three main types:
\begin{inparaenum}[(i)]
    \item decoder-only models, such as the GPT ~\cite{GPT} family,
    \item encoder-only models, such as BERT ~\cite{BERT}, and
    \item encoder-decoder models, such as T5 ~\cite{T5}.
\end{inparaenum}
Each architecture is different and works for different types of tasks. Encoder-only models, such as BERT, process the entire input sequence at once, in this case all columns and rows. After, it uses bidirectional attention, which means it tries to learn the relationships of the entire table at once ~\cite{BERT}. 

Decoder-only models, handle tabular data like a sequence. They generate one column value at a time, and use causal attention, which means each prediction only sees what came before. This is ideal for tabular data, where the model learns the step-by-step pattern of how real rows are structured ~\cite{GPT}. Encoder-decoder models, like T5, include both parts. First, the encoder reads and understands the input table. Then, the decoder takes that understanding and generates new values or a transformed version of the table ~\cite{T5}. However, this type of LLM has not been implemented for TDS.

For this study, we focused on decoder-only LLMs, which are commonly used in TDS tools. All selected models were sourced from the Hugging Face Model Hub ~\cite{hugging}, which provides a a platform for accessing and deploying pretrained Transformer models. The LLMs selected for this sensitivity analysis were chosen to cover a wide range of sizes and configurations. The final selection is shown in Table \ref{tab:LLMs}. 

\begin{table}[h]
\centering
\caption{Selected LLMs for the sensitivity assessment and their standard configurations.}
\label{tab:LLMs}
\scalebox{0.85}{
\begin{tabular}{lcccc}
\toprule
\shortstack{\textbf{LLM}} 
& \shortstack{\textbf{Standard} \\ \textbf{Layers}} 
& \shortstack{\textbf{Hidden} \\ \textbf{Dimension}} 
& \shortstack{\textbf{Attention} \\ \textbf{Heads}} 
& \shortstack{\textbf{Standard} \\ \textbf{Parameters}} \\
\midrule
GPT-2 ~\cite{GPT}            & 12 & 768   & 12 & 124M  \\
LLaMA ~\cite{llama}          & 32 & 4096  & 32 & 7B    \\
GPT-Neo ~\cite{gpt-neo}      & 32 & 2048  & 16 & 2.7B  \\
GPT-BigCode ~\cite{gpt-bigcode} & 40 & 5120  & 40 & 15B  \\
\bottomrule
\end{tabular}
}
\end{table}

\subsection{LLM Size Estimation}
\label{sec:capacity}
To ensure a fair comparison across different LLMs, we estimated the model size in each experiment, defined as the number of trainable parameters. With a single model family (e.g., GPT-2), this would be straightforward, for example, reducing the number of layers from 12 to 6 approximately cuts the parameter count in half. However, across different LLMs, the number of parameters also depends on the hidden dimension, number of attention heads, and feed-forward network size. Therefore, the first step in our method was to develop a procedure for consistently estimating LLM size across experiments.

In a transformer, the hidden dimension $H$ is the size of the internal token representation. During self-attention, the input of size $H$ is projected into queries, keys, and values. Attention scores (query $\times$ key) weight the values, and each head outputs a vector of size $H/\text{heads}$. The head outputs are concatenated and projected back to size $H$, so the hidden dimension is preserved before and after attention \cite{vaswani_attention_2017}.

A layer is one complete block of computation: multi-head self-attention followed by a feed-forward network. Each layer maps vectors of size $H$ back to $H$, and stacking many layers refines the token representations, enabling deeper reasoning and more complex pattern extraction.

This is the reasoning we use to estimate the size of an LLM: the number of parameters grows roughly with the square of the hidden dimension ($H^2$) and linearly with the number of layers ($L$). The quadratic dependence on $H$ comes from the fact that most of the parameters are in weight matrices that map vectors of size $H$ to other vectors of size $H$. Each such matrix has $H \times H$ entries, so doubling the hidden dimension multiplies the parameter count by four. The linear dependence on $L$ comes from stacking more layers, since each layer contributes approximately the same number of parameters. Therefore, both the width of the model (hidden dimension) and its depth (layers) directly determine its total parameter count, and thus its overall size.  

\begin{equation}
\label{eq:cap}
\text{Size}_{LLM} \;\approx\; c \cdot L \cdot H^2
\end{equation}

where the constant $c$ changes for each LLM family. In order to find constant $c$ for each LLM family, we used the fact that we know the number of parameters of the standard LLMs, their number of layers and hidden size. Table \ref{tab:experiment} shows constant $c$ for each LLM family and the estimated size of the LLM for each experiment.

\begin{table}[h]
\centering
\caption{Experiment configurations, showing the LLM family, the number of layers, hidden dimension, best constant $c$, and estimated LLM size.}
\label{tab:experiment}
\scalebox{0.8}{
\begin{tabular}{lcccc}
\toprule
\textbf{LLM} & \textbf{Experiment Layers ($L$)} & \textbf{Hidden Dimension ($H$)} & \textbf{constant ($c$)} & \textbf{LLM Size ($Size_{LLM}$)} \\
\midrule
GPT-2            & 6  & 768   & 18 & 57M  \\
GPT-2            & 12 & 768   & 18 & 113M \\
GPT-Neox         & 1  & 2048  & 12 & 528M \\
GPT-Neo          & 2  & 2048  & 20 & 151M \\
GPT-Neo          & 4  & 2048  & 20 & 302M \\
GPT-Neo          & 6  & 2048  & 20 & 453M \\
GPT-Neo          & 8  & 2048  & 20 & 604M \\
GPT-J            & 1  & 4096  & 13 & 218M \\
GPT-BigCode      & 12 & 5120  & 14 & 3.46B \\
GPT-BigCode      & 6  & 5120  & 14 & 1.73B \\
LLaMA            & 2  & 4096  & 13 & 403M \\
LLaMA            & 1  & 4096  & 13 & 201M \\
\bottomrule
\end{tabular}
}
\end{table}

\subsection{Datasets}
\label{sec:datasets}
For the evaluation, we selected four widely used tabular datasets: \textit{adult} \cite{uci_adult}, \textit{customer} \cite{customer}, \textit{house} \cite{kaggle_house}, and \textit{stroke\_healthcare} \cite{stroke}. The \textit{adult} dataset predicts income level by classifying salaries under and over \$50.000, the \textit{customer} dataset models whether customers will leave the bank, the \textit{house} dataset predicts median house values, and the \textit{stroke\_healthcare} dataset estimates stroke risk from patient records. These datasets provide a mix of demographic, financial, and healthcare use cases, covering both classification and regression tasks. The dataset characteristics are shown in Table \ref{tab:datasets}.

\begin{table}[h]
\centering
\caption{Characteristics of the datasets used in this study.}
\label{tab:datasets}
\scalebox{1}{
\begin{tabular}{lccccc}
\toprule
\textbf{Dataset} & \textbf{Rows} & \textbf{Columns} & \textbf{Categorical} & \textbf{Continuous} & \textbf{Task} \\
\midrule
Stroke\_healthcare \cite{stroke} & 5110  & 11 & 7 & 4 & Classification \\
Customer  \cite{customer}          & 6400  & 22 & 16 & 6 & Classification \\
House  \cite{kaggle_house}             & 13690 & 9  & 1 & 8 & Regression \\
Adult  \cite{uci_adult}             & 30260 & 14 & 9 & 5 & Classification \\
\bottomrule
\end{tabular}
}
\end{table}

We preprocessed the datasets to ensure consistent evaluation: 
\begin{inparaenum}[(i)]
    \item Categorical columns were converted into numerical representations using encoding, 
    \item Continuous columns were normalized to reduce the impact of scale differences and improve comparability, 
    \item Missing values were handled by removing any incomplete records, and 
    \item All datasets were shuffled and split to avoid any temporal or positional bias.
\end{inparaenum}

All experiments were conducted on a Linux laptop with 32GB RAM, an Intel Core i9-12900H, and an external NVIDIA RTX 4090 GPU (24GB VRAM) in a Razer Core X eGPU enclosure with a 1000W PSU.

\section{Results}
\label{sec:results}
This section presents the results of the sensitivity analysis along the three evaluation dimensions: runtime, ML utility, and similarity. Each dimension captures a different aspect of the TDS tool's behavior and together they provide a comprehensive view of the trade-offs between efficiency, predictive performance, and data fidelity. The following subsections present the detailed results for each dimension and highlight the main patterns observed across datasets, tools, and LLM configurations.

\subsection{Runtime}
\label{res:runtime}
The first aspect examined in this sensitivity analysis is runtime, defined as the total time required for training and synthetic data generation. Runtime was measured in seconds using a monitoring shell script on Linux. To increase the reliability of the results, each experiment was repeated five times, and the average runtime was reported.

\begin{figure}[h]
    \centering
    \includegraphics[width=1\linewidth]{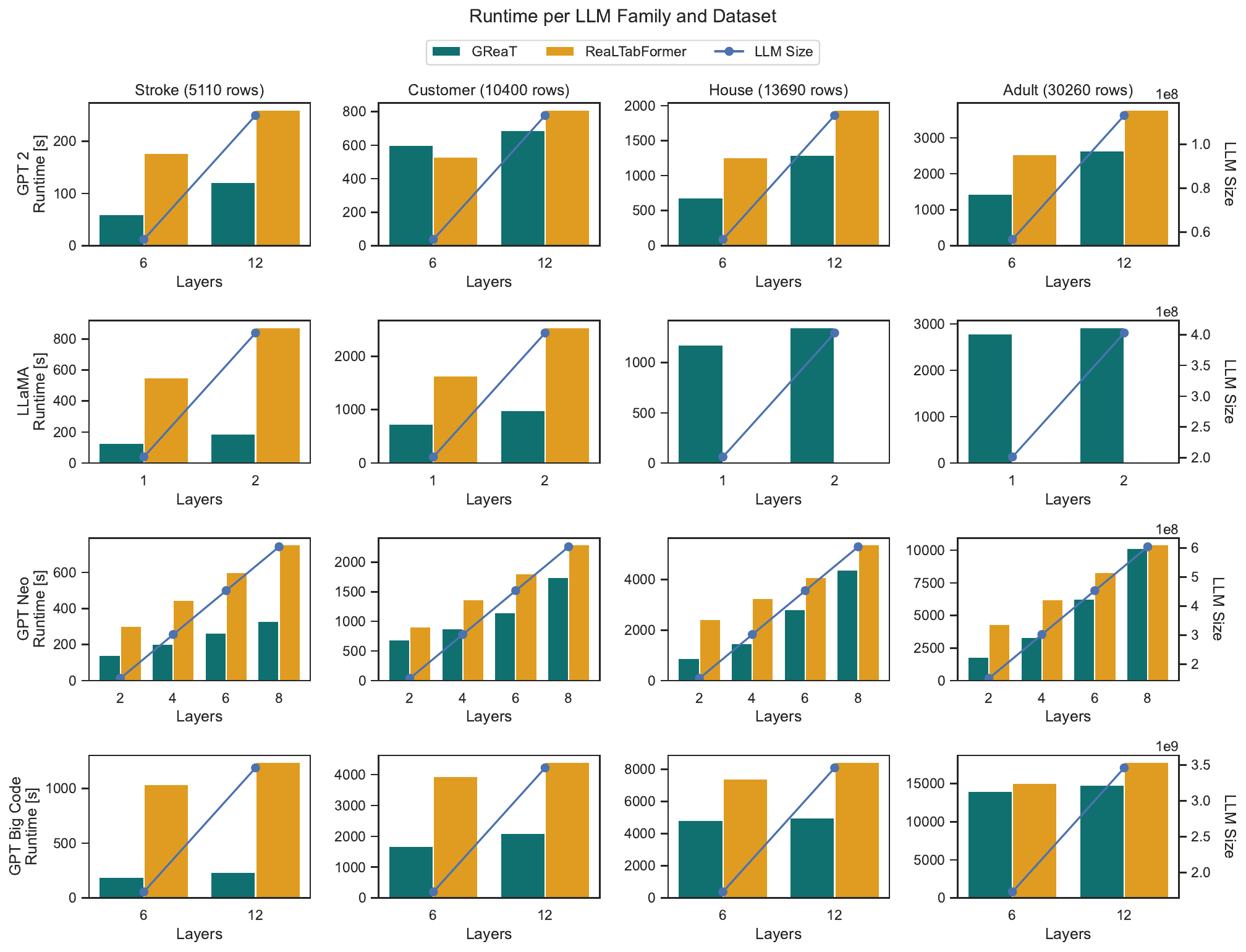}
    \caption{Runtime comparison across LLM families and datasets. Bars show runtime (in seconds) per number of layers and tool (GReaT vs. ReaLTabFormer). The overlaid line indicates the corresponding LLM size (parameters), plotted on the secondary y-axis.}
    \label{fig:runtime}
\end{figure}

The results are shown in Figure~\ref{fig:runtime}, which presents a separate plot for each dataset and LLM. Results are reported for both tools, with ReaLTabFormer shown in orange and GReaT in teal. Each plot combines two components: the runtime, represented as bars, and the size of the LLM, represented as a line.

The runtime results in Figure~\ref{fig:runtime} show consistent trends across all datasets that highlight the impact of both the choice of TDS tool and the LLM configuration. An initial observation is that ReaLTabFormer did not converge on the larger datasets when using the LLaMA model.

Overall, GReaT consistently outperforms ReaLTabFormer in terms of runtime, often requiring substantially less time for the same number of layers. The advantage is most pronounced on smaller datasets, such as \textit{stroke} and \textit{customer}, while the gap narrows as dataset size increases. For larger datasets like \textit{house} and particularly \textit{adult}, runtimes for ReaLTabFormer and GReaT begin to converge. This suggests that the relative efficiency of GReaT holds primarily for small to medium-sized datasets. 

Another important factor is the number of layers in the LLM configuration, which has a direct and systematic effect on runtime across all datasets and tools. This finding confirms the initial claim of this work that runtime improves with fewer layers. The pattern is particularly clear for ReaLTabFormer, where runtime 
consistently increases with the number of layers regardless of the dataset size. 

In contrast, for GReaT this effect depends on dataset size. On the smaller datasets, runtime grows only marginally with LLM size, suggesting that the tool is less sensitive to model depth. Still, for both tools the trend remains that shallow configurations, such as the two-layer versions of GPT Neo or LLaMA, complete substantially faster than deeper configurations, while models with six or eight layers require significantly more runtime. This scaling effect is visible in every dataset: runtime grows almost linearly with the number of layers for smaller models, and accelerates further in larger configurations such as GPT Big Code. The question is whether these few-layer configurations are still able to capture the patterns in the dataset, or whether this comes at the cost of reduced utility and similarity.
  
While the absolute runtimes differ depending on the dataset size, 
the relative ordering of models and tools is preserved: shallow configurations of GPT 2, and LLaMA are always the fastest, whereas GPT Big Code configurations, particularly with 12 layers, are consistently the slowest.  

In short, the results illustrate two central drivers of runtime: 
\begin{inparaenum}[(i)]
    \item the efficiency of the underlying tool, with GReaT consistently outperforming ReaLTabFormer, especially on smaller datasets, while this advantage decreases as dataset size grows and runtimes begin to converge on larger datasets, and 
    \item the depth of the LLM, with runtime improving for shallower models but increasing substantially with the number of layers. This scaling effect is visible across all datasets: shallow configurations of GPT 2, and LLaMA, are always the fastest, whereas deeper models, particularly GPT Big Code with 12 layers, are consistently the slowest. 
\end{inparaenum}

\subsection{ML Utility}
\label{res:utility}
The second evaluation in the LLM selection sensitivity analysis focused on \textbf{ML utility}. Here, ML models were trained on both the original datasets and the synthetic datasets generated by GReaT and ReaLTabFormer. Their performance was assessed using standard metrics, which were then compared between real and synthetic training. The goal was to determine whether models trained on synthetic data could achieve performance comparable to, or better than, those trained on real data. ML models and evaluation metrics used are shown in Table \ref{tab:ml_models_metrics}.

\begin{table}[h]
\centering
\caption{ML models and evaluation metrics used for classification and regression tasks.}
\label{tab:ml_models_metrics}
\scalebox{0.9}{
\begin{tabular}{lcc}
\toprule
\textbf{Task} & \textbf{ML Models} & \textbf{Evaluation Metrics} \\
\midrule
Classification & Logistic Regression, Random Forest & Accuracy, Macro-F1 \\
Regression     & Linear Regression, Random Forest   & $R^2$ \\
\bottomrule
\end{tabular}
}
\end{table}

\begin{figure}
    \centering
    \includegraphics[width=1\linewidth]{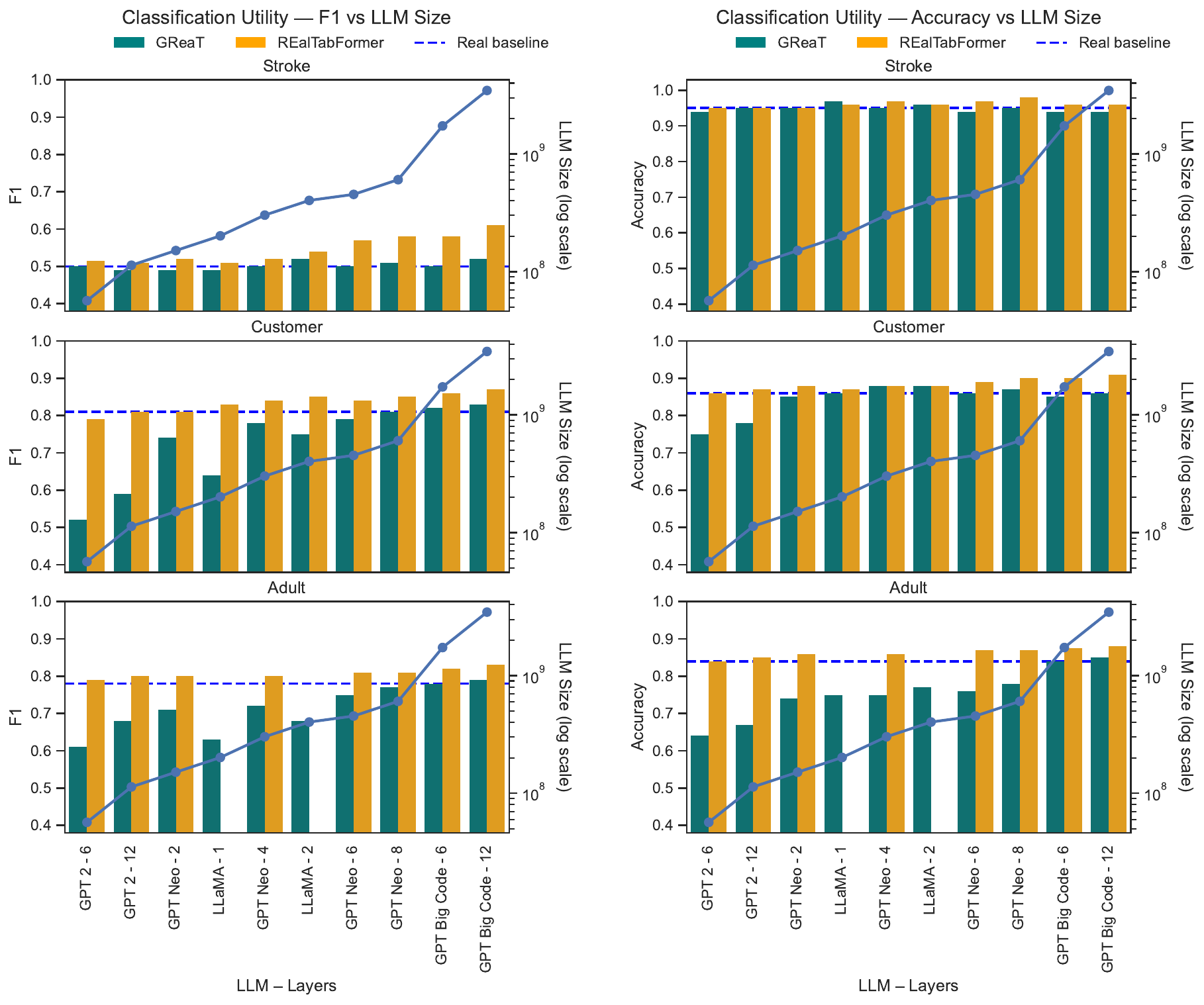}
    \caption{Accuracy and F1 of the classification task (Stroke, Customer, Adult datasets). The real-data baseline is included for reference. A secondary axis shows the LLM size on a log scale starting at $10^7$.}
    \label{fig:classification}
\end{figure}

\begin{figure}
    \centering
    \includegraphics[width=0.8\linewidth]{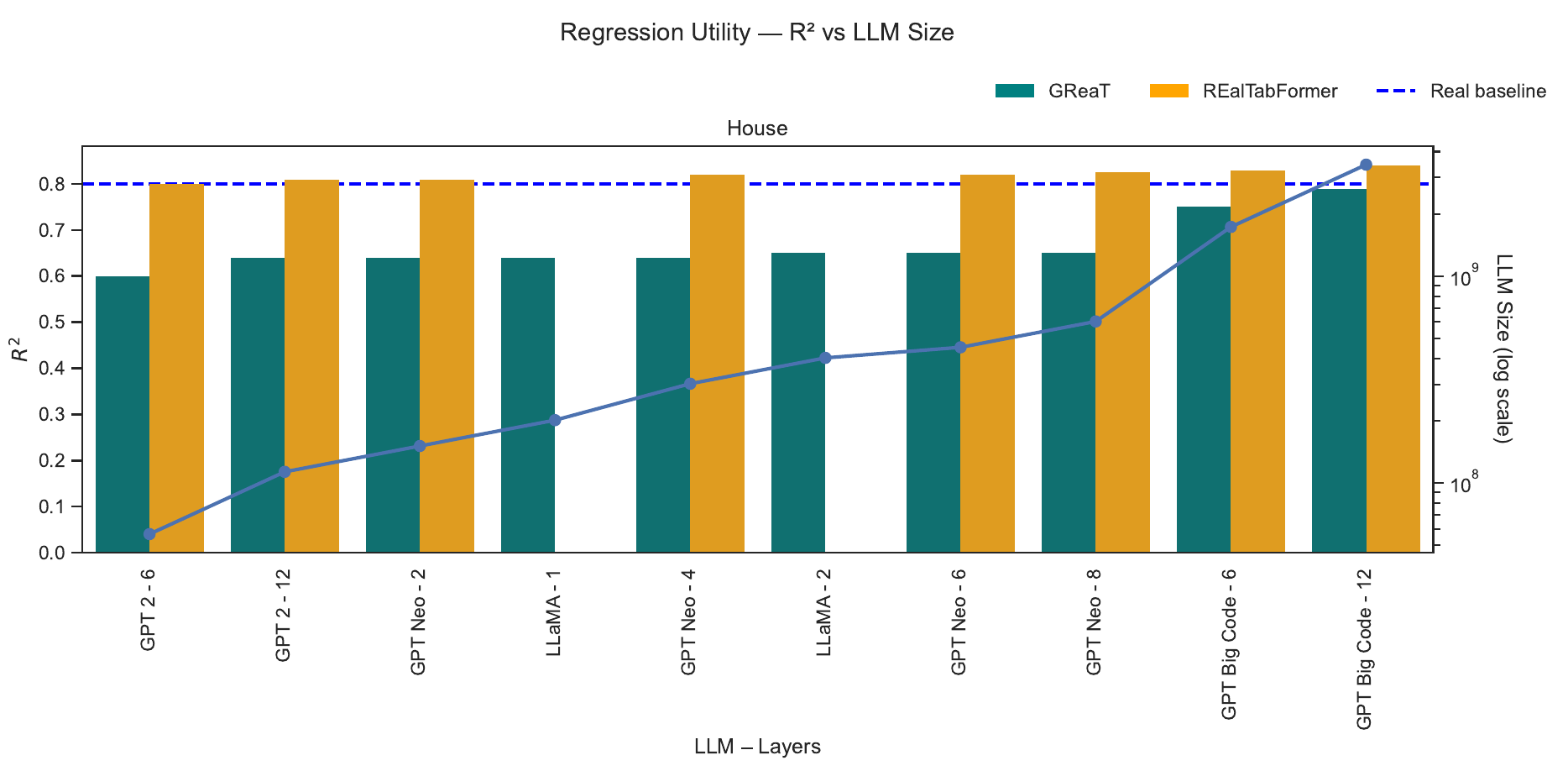}
    \caption{\(R^2\) of the regression task (House dataset). The real-data baseline is included. A secondary axis shows the LLM size on a log scale.}
    \label{fig:utility_regression}
\end{figure}

Each experiment with each ML model was run five times to increase reliability. We report the mean score per dataset. The results are shown in Figures~\ref{fig:classification}, 
and~\ref{fig:utility_regression}. For classification, we plot F1 and Accuracy, for regression, we plot \(R^2\). The bars represent the metrics based on the synthetic data from GReaT and ReaLTabFormer with the different LLMs. The x-axis lists the LLM model and layer count used to generate the synthetic data, and a secondary y-axis shows the capacity on a logarithmic scale.

The first observation regarding ML utility is that, for the smallest dataset (\textit{Stroke}), synthetic data from both TDS tools reach a utility level comparable to the original data. In fact, REaLTabFormer even outperforms the baseline on this task. This is a good and expected result, since one of the goals of synthetic data is to keep or improve model performance when the real dataset is very small. For the other three datasets, however, synthetic data generated with GReaT consistently underperforms compared to the original data, showing a clear loss in utility.

When comparing configurations within the same LLM family, REaLTabFormer shows stable ML utility across layer reductions, suggesting that fewer layers do not decrease downstream performance. In contrast, the utility of the datasets generated with GReaT declines as the number of layers decreases.

\subsection{Similarity}
\label{res:similarity}
The final evaluation aspect in the LLM sensitivity analysis is similarity. To evaluate how similar the synthetic data is to the real data, a classifier-based approach was used. Specifically, a Random Forest model was trained as a Discriminator to distinguish between real and synthetic samples. The model was given a binary classification task where the input data combined both real and synthetic records, each labeled accordingly.

\begin{figure}[h]
    \centering
    \includegraphics[width=1\linewidth]{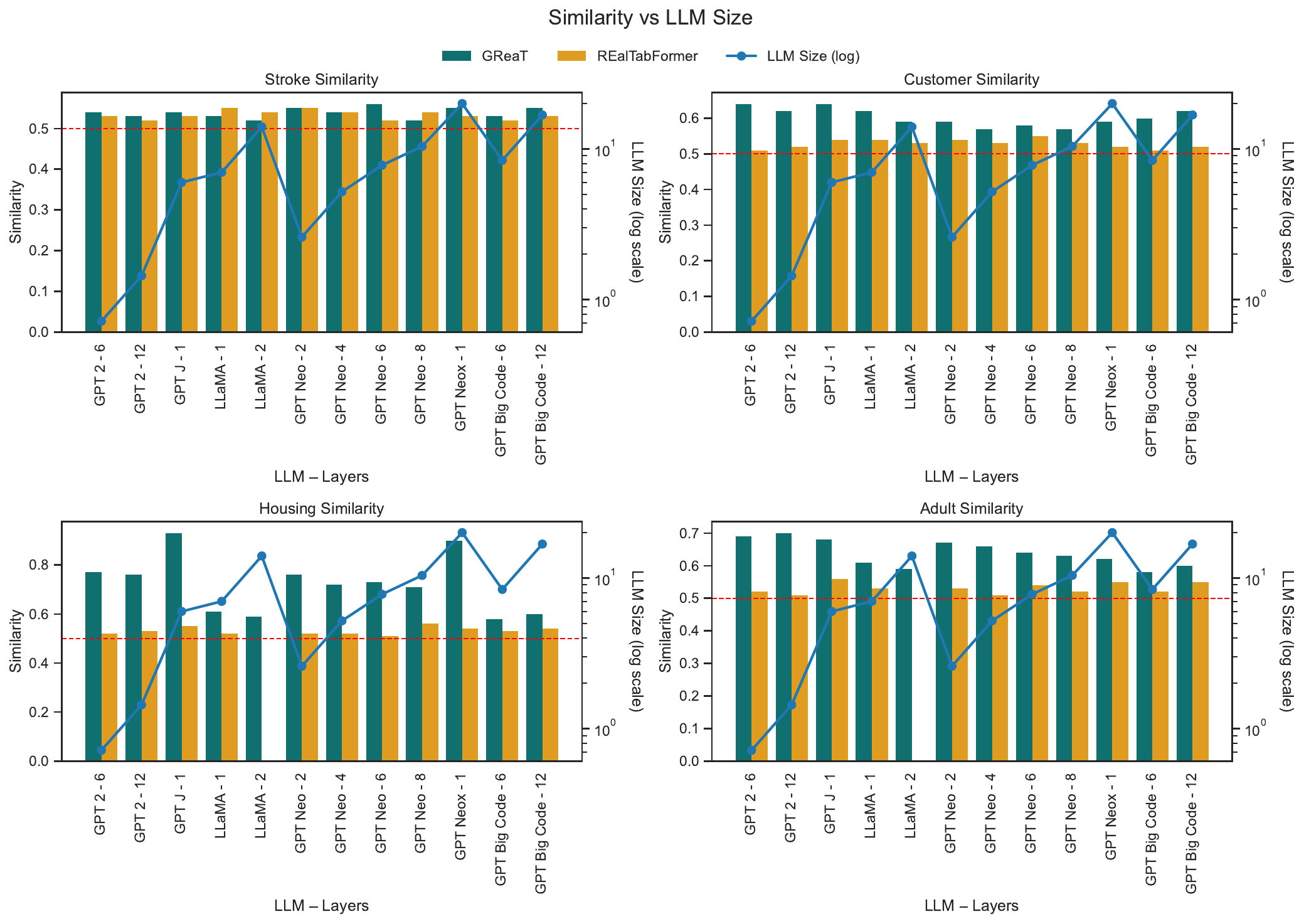}
    \caption{Similarity between real and synthetic datasets for each LLM configuration and TDS tool across the four evaluation datasets. Bars represent similarity scores, with the red dashed line marking the 0.5 threshold. The black line indicates model capacity on a logarithmic scale.}
    \label{fig:llm_similarity}
\end{figure}

The accuracy of the Discriminator is a measure of similarity between the two distributions. If the classifier achieves high accuracy, it means the real and synthetic data are easily distinguishable, indicating low similarity between real and synthetic distributions. In contrast, if the accuracy is close to 50\%, the classifier performs no better than random guessing, suggesting that the synthetic data is statistically similar to the real data. This metric does not assess privacy or utility directly but provides an indication of how realistically the synthetic data mimics the original distribution. The accuracy results for the similarity evaluation is shown in Figure \ref{fig:llm_similarity}.

Just like with ML utility, for the smallest dataset \textit{Stroke} both TDS tools achieve the ideal result, where the synthetic datasets have similarity close to 0.5, which means the discriminator is having a hard time identifying the synthetic from the real one. This indicates that for this dataset, both GReaT and REaLTabFormer generate data that closely mimics the original distribution. For REaLTabFormer, this pattern continues across all datasets, the discriminator’s accuracy remains close to the 0.5 threshold, suggesting that the generated datasets are consistently hard to distinguish from the real ones, regardless of the LLM configuration or dataset size.

In contrast, the results for GReaT diverge from this trend. For the larger datasets, the discriminator achieves substantially higher accuracy, meaning it can more easily separate synthetic from real records. This indicates that GReaT’s synthetic data departs more strongly from the real distribution as dataset complexity increases. The gap between the two tools becomes particularly visible in datasets such as \textit{Adult} and \textit{Customer}, where REaLTabFormer remains close to the ideal similarity value, while GReaT outputs synthetic datasets that are easily identified as artificial.

\section{Conclusion}
\label{sec:conclusion}

The goal of this work was to assess whether it is possible to reduce the computational requirements of Transformer-based TDS tools while maintaining data utility and similarity to real distributions. We carried out this sensitivity analysis by systematically varying the choice of LLMs across two widely used tools, GReaT and REaLTabFormer, and evaluating them along three dimensions: runtime, ML utility, and similarity.

Our findings include three main insights. First, runtime scaled as expected, with shallower LLMs consistently performing better. GReaT achieved lower runtimes than REaLTabFormer, though the difference became smaller on the larger dataset. However, this efficiency came at the expense of utility and similarity, limiting the benefits of using shallow LLMs with GReaT.

With respect to data quality, we observed that for small datasets both TDS tools behaved as expected, where the ML utility of 
the synthetic data remained similar or better compared to the real dataset, and the synthetic records could not be easily distinguished from real ones, indicating optimal similarity. For larger datasets, however, only REaLTabFormer maintained high utility and similarity values close to 0.5, whereas GReaT diverged significantly.

As a general conclusion, only REaLTabFormer performs appropriately with simple LLM configurations, showing that the quality of synthetic datasets is not compromised even when using lightweight models with substantially lower runtime than larger ones. Nevertheless, this runtime remains higher compared to GReaT and other TDS tools, 
suggesting that computational performance can be improved, but only up to a certain level.

While our study provides valuable insights into the sensitivity of Transformer-based TDS tools, it also has several limitations that represent our future work. First, we restricted our evaluation to two Transformer-based tools (GReaT and REaLTabFormer) and a selected set of LLM families. It remains unclear whether the observed trade-offs generalize to other Transformers. Second, our experiments used four datasets. Although these cover both regression and classification tasks, they do not reflect very large-scale, high-dimensional, or domain-specific datasets. Extending the evaluation to larger and more heterogeneous datasets would strengthen the generalizability of our findings. Finally, our analysis focused on runtime, ML utility, and similarity to real data distributions, but did not explicitly assess privacy preservation, which is a central motivation for synthetic data. Future work should incorporate privacy risk evaluation to ensure that efficiency gains do not come at the expense of sensitive data leakage.

\bibliographystyle{unsrt}  
\bibliography{references}

\end{document}